\documentclass[
  10pt,
  logo,
  onecolumn,
  copyright,
  nodate,
  colorlinks=true,
  allcolors=blue
]{nvidiatechreport}

\usepackage[numbers,sort,square]{natbib}
\usepackage{flafter}

\renewcommand*{\backref}[1]{}
\renewcommand*{\backrefalt}[4]{}
\renewcommand{\headingfont}{\color{black}\bfseries\fontsize{12}{13}\selectfont}

\titlespacing*{\section}{0pc}{2.481875ex plus3.61pt minus2.7075pt}{5pt}
\titlespacing*{\subsection}{0pc}{2.07575ex plus2.7075pt minus1.805pt}{2pt}

\title{Effective Teacher Support for Reasoning Distillation}
\author{\large Lavinia Ghita, Dhruv Desai\\
\normalfont NVIDIA}
\date{}

\begin{document}

\maketitle

\noindent\textbf{Abstract.}\enspace
Forward-KL language-model distillation normally computes dense teacher and
student scores over the full vocabulary at every supervised position, making
training and stored teacher targets expensive. We instead distill from the
teacher's highest-probability vocabulary fraction, renormalizing the retained
mass as we sweep support from 100\% to 0.01\%. Our main study distills Qwen3
4B\(\rightarrow\)1.7B. We confirm the broader support-reduction finding on
larger Qwen3 8B\(\rightarrow\)4B and Nemotron Nano
12B\(\rightarrow\)9B pairs and under answer-only supervision. We evaluate
generation primarily on MMLU-Pro and secondarily on GPQA-Diamond. Across the
Qwen3 settings, retaining 0.01\% of the vocabulary causes no observed
MMLU-Pro degradation relative to full-vocabulary distillation, although
disagreement with the complete teacher distribution increases. Nemotron Nano
matches full-vocabulary MMLU-Pro at 0.1\%, but smaller fractions fail to emit
the closing reasoning tag, showing that the reliable cutoff depends on the
model family. At 0.01\% support, caching the retained next-token scores from
the teacher forward pass cuts training step time by 38\%, while restricting
student output computation to those tokens cuts peak memory by 24\%; neither
causes observed MMLU-Pro degradation.
\par

\section*{Introduction}

Knowledge distillation trains a student to match a teacher's predictive
distribution, so that a smaller model can inherit the generalization of
a larger one~\cite{Hinton:2015aa}. For an autoregressive language model,
this predictive distribution is the next-token softmax over the vocabulary.
Token-level LLM distillation commonly matches this vocabulary-wide
distribution and evaluates the resulting student by generation
quality~\cite{Gu:2023aa,Agarwal:2024gkd,Ko:2024aa}.

Standard next-token cross-entropy fine-tuning uses a one-hot target, whose
support is one, at each supervised position, even though the student softmax
remains normalized over the full vocabulary. This target sparsity motivates
asking whether distillation likewise requires full-vocabulary teacher
targets. This analogy is imperfect, however, because a teacher's relative
probabilities over alternative tokens can carry useful information.

Teacher probabilities can nevertheless be highly concentrated, making
sparse targets appealing at LLM scale~\cite{Shum:2024first}. Yet high
retained mass alone does not rule out degradation under a truncated
objective~\cite{Anshumann:2025sls}. How far support can be reduced without
observed generation degradation is therefore an empirical question.

We study this question for distillation on reasoning
traces~\cite{Wei:2022aa}. We first measure the effective support of the
frozen teacher, then train students with rank-\(k\) truncated forward KL by
retaining its \(k\) highest-probability tokens at each supervised position.
We compare these students with an otherwise identical full-vocabulary
student on MMLU-Pro~\cite{MMLU-Pro:2024} and
GPQA-Diamond~\cite{Rein:2024gpqa}. To distinguish optimization of each
truncated objective from agreement with the complete teacher distribution,
we report both the \(k\)-dependent training loss and a common
full-vocabulary KL diagnostic.

Our study provides a controlled empirical account of retained teacher
support in reasoning distillation. We characterize that support, test a
rank-\(k\) ladder against full-vocabulary KL, and relate the resulting loss
diagnostics to generation. We also probe a second training setting with a
separate answer-only experiment, confirm the support-size result on a larger
Qwen3 teacher--student pair (Qwen3-4B from Qwen3-8B)~\cite{Qwen3:2025}, and
test cross-family transfer on an auxiliary Nemotron Nano 12B\(\rightarrow\)9B
pair~\cite{NVIDIA:2025nemotronnano2}. Finally, we quantify the systems
consequences under matched configurations. Caching \(k=15\) targets (0.01\%
of the vocabulary) removes the teacher from the training loop and reduces
step time by 38\%, while compact \(k\)-row output computation avoids
materializing vocabulary-sized student logits and reduces peak allocated
memory by 24\% relative to cached full-head training. We report these systems
effects separately from generation quality and verify that both cached
variants show no observed MMLU-Pro degradation.

The remainder of the paper is organized as follows.
Section~\ref{sec:motivation} motivates effective teacher support, and
Section~\ref{sec:related} reviews the relevant distillation literature.
Section~\ref{sec:method} defines the retained-support objectives.
Section~\ref{sec:setup} describes the data and experimental protocol, and
Section~\ref{sec:results} presents the results.
Section~\ref{sec:conclusion} concludes the paper, and
Appendices~\ref{app:truncation} and~\ref{app:gumbel} decompose the
retained-support objectives and derive the teacher-sampling construction.

\section{Motivation}
\label{sec:motivation}

At each supervised position \(t\), forward-KL distillation compares a
teacher distribution \(p_t\) and a student distribution \(q_t\) over the
vocabulary of size \(V\):
\begin{equation}
\label{eq:full-kl}
  \mathcal{L}_t
  = \sum_{v=1}^{V} p_t(v)
    \bigl(\log p_t(v)-\log q_t(v)\bigr).
\end{equation}
The softmax assigns positive probability to every token, so its mathematical
support is all of \(V\), even when most of its mass is concentrated on a few
tokens. If \(p_{t,(i)}\) is the \(i\)-th largest teacher probability and
\(0<\rho\leq1\), define
\[
  k_\rho(t)
  = \min\left\{k:\sum_{i=1}^{k}p_{t,(i)}\geq\rho\right\}.
\]
This is the smallest rank cutoff that retains at least a fraction \(\rho\)
of the teacher mass at position \(t\). Its empirical distribution is over
positions, not token identities or individual probabilities, and maps a
target mass to a candidate truncation scale.

FIRST~\cite{Shum:2024first} shows that a few teacher tokens can retain most
probability mass at LLM scale. Minitron and Sparse Logit Sampling show
that aggressive top-\(k\) objectives can nevertheless degrade the student
relative to full-distribution distillation~\cite{Muralidharan:2024minitron,
Anshumann:2025sls}. We therefore use \(k_\rho\) to identify candidate
support sizes, then test for observed generation degradation.

The distillation temperature \(\tau\) scales the teacher logits before the
softmax; \(\tau=1\) leaves the native next-token distribution
unchanged~\cite{Hinton:2015aa}.
Figure~\ref{fig:teacher-support} reports empirical CDFs of \(k_\rho\) for
\(\rho\in\{0.90,0.95,0.99\}\) for frozen Qwen3-4B~\cite{Qwen3:2025} at
\(\tau=1\). At \(k=15\), the retained tokens contain at least 90\%, 95\%, and
99\% of the teacher mass at 99.0\%, 97.9\%, and 93.4\% of later supervised
positions, respectively. The first supervised position has \(k_\rho=1\) in
every example and is omitted. These CDFs use the native teacher ranking,
before the Qwen3-specific tag correction applied during training in
Section~\ref{sec:method}.

\begin{center}
\centering
\includegraphics[width=0.7003\linewidth]{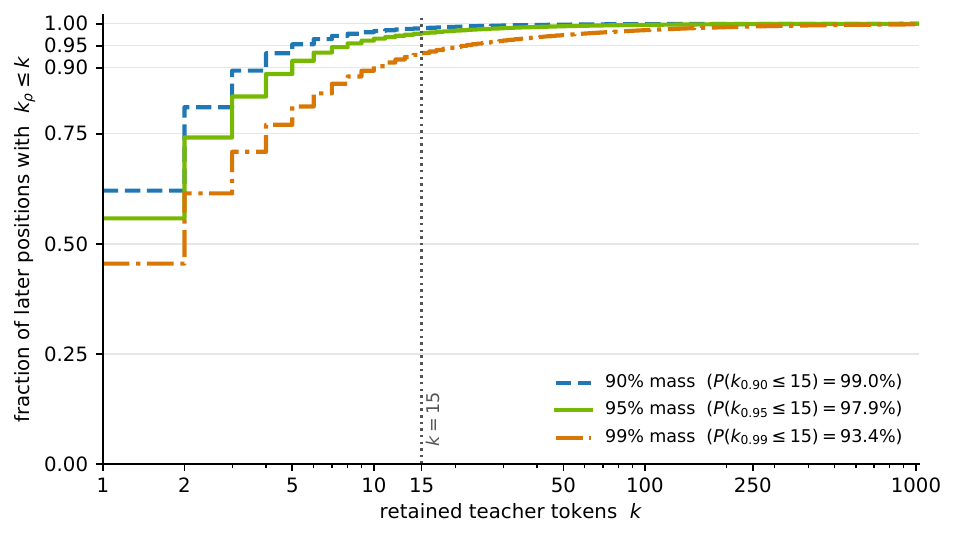}
\captionof{figure}{\textbf{Effective teacher support on reasoning traces.}
Empirical CDFs of \(k_\rho\) over the \(905{,}193{,}844\) supervised
positions remaining after excluding the first supervised position in each
example, for which \(k_\rho=1\). The vertical line marks \(k=15\).}
\label{fig:teacher-support}
\end{center}

\noindent
The reasoning ladder extends to 0.01\% of the 151,936-token vocabulary,
corresponding to \(k=15\). Figure~\ref{fig:teacher-support} shows that this
extreme truncation still retains most of the teacher mass at most positions,
but retained mass alone does not establish that generation will be
unaffected. Training optimizes a token-level objective under teacher-forced
contexts, whereas benchmark evaluation requires the student to generate
complete solutions autoregressively. We therefore compare rank-\(k\) and
full-vocabulary students on MMLU-Pro and GPQA-Diamond, and place their
\(k\)-dependent training losses alongside a common full-vocabulary KL
diagnostic. A separate answer-only experiment examines a second training
setting (Section~\ref{sec:answer-only}).

For offline distillation, a dense teacher-target cache stores \(O(V)\)
values per supervised position, whereas a rank-\(k\) cache stores \(O(k)\)
token identifiers and values. Caching removes the teacher from the training
loop. For the renormalized objective, the loss also requires only the
selected student logits, making restricted output computation possible
(Section~\ref{sec:offline}).

\section{Related work}
\label{sec:related}

Classical knowledge distillation trains a student on a teacher's softened
predictive distribution, whose low-probability classes can encode useful
information about the teacher's generalization behavior. Temperature-based
distillation flattens the teacher distribution and gives lower-probability
classes greater influence~\cite{Hinton:2015aa}. Our findings point in a
different direction in this setting: at \(\tau=1\), generation on the primary
benchmark remains stable even when the native teacher tail is omitted. This
suggests that further emphasizing the tail through a higher distillation
temperature may be unnecessary here.

Modern LLM distillation changes either the divergence or the distribution of
training sequences. MiniLLM optimizes reverse KL on student-generated
responses~\cite{Gu:2023aa}, Generalized Knowledge Distillation combines
fixed and student-generated sequences with several divergence
choices~\cite{Agarwal:2024gkd}, and DistiLLM introduces skew KL with
adaptive off-policy training~\cite{Ko:2024aa}. These methods alter the
divergence or training contexts rather than selecting a fixed vocabulary
subset for forward-KL supervision. Reasoning distillation commonly transfers
teacher-generated rationales as sequence supervision~\cite{Ho:2023reasoning};
our study instead examines the effective vocabulary support of the teacher
distribution at each position within those traces. Complementary work
reduces reasoning-distillation cost by truncating the supervised sequence
rather than the vocabulary axis~\cite{Chen:2026essence}.

Several methods reduce the teacher output at LLM scale. SLIM stores the top
5\% of teacher logits~\cite{Raman:2023slim}. FIRST reports that five tokens
can contain more than 95\% of the probability mass and distills from
recalibrated top-five targets~\cite{Shum:2024first}. A design-space study of
pre-training distillation explores offline top-\(p\) and top-\(k\) targets
for LLM pretraining~\cite{HaoPeng:2025pretraining}.
High retained mass alone does not rule out degradation from a truncated
objective. Minitron reports degradation for small top-\(k\) objectives in
post-pruning recovery~\cite{Muralidharan:2024minitron}, while Sparse Logit
Sampling analyzes the bias introduced by truncation and proposes an
importance-sampled estimator that preserves the full-KD gradient in
expectation~\cite{Anshumann:2025sls}. RLKD~\cite{Peng:2025rlkd} is
complementary: it aligns the ranking of teacher and student peaks rather
than restricting forward KL to the teacher's retained support.

Multi-axis selective KD~\cite{Tavor:2026selective} jointly selects positions,
vocabulary classes, and training samples. It uses importance-sampled class
selection and reports reductions in storage, wall time, and peak memory. A
fused, chunked offline-KL implementation~\cite{Ryskulov:2026offline}
directly caches partial-mass top-100 targets while retaining the
full-vocabulary student normalizer. Our study instead isolates deterministic
teacher-ranked support and selects \(k\) through generation on reasoning
traces. We then apply the same offline caching pattern at \(k=15\). Our
teacher-sampled robustness check has no importance correction and is
therefore not an unbiased estimator of full KL~\cite{Anshumann:2025sls}.
The compact renormalized \(k\)-row path is technically distinct because it
removes the full-vocabulary student normalizer; we treat it as a secondary
systems path.

\section{Method}
\label{sec:method}

\subsection{Retained-support objectives}

Let \(z_t^{\mathrm T},z_t^{\mathrm S}\in\mathbb{R}^{V}\) denote the teacher
and student logits at supervised position \(t\). At distillation temperature
\(\tau>0\), define
\[
  p_t^\tau=\operatorname{softmax}(z_t^{\mathrm T}/\tau),
  \qquad
  q_t^\tau=\operatorname{softmax}(z_t^{\mathrm S}/\tau).
\]
Equation~\eqref{eq:full-kl} is the full-vocabulary forward KL at \(\tau=1\).
For general \(\tau\), we use \(p_t^\tau,q_t^\tau\) and multiply the loss by
\(\tau^2\), following classical knowledge
distillation~\cite{Hinton:2015aa}. We fix \(\tau=1\), measuring support on
the teacher's native next-token distribution; \(\tau\) is distinct from the
decoding temperature used for generation.

At each position, let \(\pi_t\) be a permutation of the token identifiers
that orders the teacher logits from largest to smallest, with ties ordered
arbitrarily:
\[
  \begin{aligned}
  z_t^{\mathrm T}\!\left(\pi_t(1)\right)
  &\geq \cdots \geq
  z_t^{\mathrm T}\!\left(\pi_t(V)\right),\\
  S_t&=\{\pi_t(1),\ldots,\pi_t(k)\},
  \qquad
  M_t=\sum_{v\in S_t}p_t^\tau(v).
  \end{aligned}
\]
Thus \(S_t\) contains the \(k\) most probable teacher tokens, since softmax
preserves the logit ordering, and \(M_t\) is their total probability mass.
The primary objective renormalizes both distributions within \(S_t\),
yielding the conditional teacher and student distributions \(\bar p_t\) and
\(\bar q_t\):
\[
  \bar p_t(v)=\frac{p_t^\tau(v)}{M_t},
  \qquad
  \bar q_t(v)=
  \frac{\exp(z_t^{\mathrm S}(v)/\tau)}
       {\sum_{u\in S_t}\exp(z_t^{\mathrm S}(u)/\tau)}.
\]
The corresponding per-position loss, denoted
\(\ell_t^{\mathrm{renorm}}\), is the forward KL between these conditional
distributions:
\[
  \ell_t^{\mathrm{renorm}}
  =\tau^2\sum_{v\in S_t}
  \bar p_t(v)\log\frac{\bar p_t(v)}{\bar q_t(v)}.
\]
This objective compares the relative probabilities within the selected set.
It removes the teacher tail and the full-vocabulary student normalizer.

We also evaluate a partial-mass objective,
\[
  \ell_t^{\mathrm{mass}}
  =\tau^2\sum_{v\in S_t}
  p_t^\tau(v)\log\frac{p_t^\tau(v)}{q_t^\tau(v)}.
\]
It retains the teacher's original mass on \(S_t\) and the full-vocabulary
student normalizer~\cite{Ryskulov:2026offline}. For a proper unmodified
subset \(S_t\), the retained teacher probabilities sum to \(M_t<1\), so this
truncated sum is not itself a KL divergence and need not be nonnegative.
Both objectives recover full-vocabulary forward KL when \(k=V\). Their
different normalizations also expose the bias introduced by
truncation~\cite{Anshumann:2025sls}. We report renormalized KL as the primary
method because it compares teacher and student distributions conditioned on
the selected support and permits student computation restricted to the
retained rows. An earlier auxiliary rank-\(k\) sweep found similar generation
behavior for the two objectives. We retain the partial-mass objective as an
objective-design comparison.
Appendix~\ref{app:truncation} gives an exact decomposition and the
corresponding student-logit gradients.

\subsection{Support construction}

Rank selection is deterministic and defines the main \(k\)-ladder. As a
robustness check, draw independent \(g_v\sim\operatorname{Gumbel}(0,1)\)
and let \(\pi_t^g\) order the perturbed teacher logits from largest to
smallest:
\[
  \frac{z_t^{\mathrm T}(\pi_t^g(1))}{\tau}+g_{\pi_t^g(1)}
  \geq \cdots \geq
  \frac{z_t^{\mathrm T}(\pi_t^g(V))}{\tau}+g_{\pi_t^g(V)},
  \qquad
  S_t^{\mathrm{sample}}
  =\{\pi_t^g(1),\ldots,\pi_t^g(k)\}.
\]
This is sampling without replacement from the teacher distribution
\cite{Kool:2019gumbel}. We apply the same truncated objective without
proposal or inclusion-probability weights, so teacher sampling is not an
importance-sampling estimator of full KL. Uniform selection draws \(k\)
distinct vocabulary identifiers uniformly and serves as a negative control.
Uniform selection produced a much worse training objective, and its
partial-mass variant diverged.
An earlier auxiliary sweep found no systematic MMLU-Pro advantage for
teacher sampling over rank selection. Because this check was not repeated
in the three-seed ladder, we treat it as qualitative evidence and exclude it
from the main results.

Qwen3 uses \texttt{<think>} and \texttt{</think>} as structural generation
tokens~\cite{Qwen3:2025}. They mark entry to and exit from reasoning, so
omitting either from a small support removes direct supervision for that
transition. Preliminary runs without explicit retention of these tags often
failed to emit \texttt{</think>}, leaving reasoning traces unclosed.
Retaining both tags improved generation stability, though it did not prevent
every failure. We therefore append missing tags, giving at most \(k+2\)
tokens. An inserted tag copies the weight of a tag already present; if
neither is present, both receive the smallest retained weight. We rescale the
corrected weights only when their total exceeds \(1\) and apply the same
correction at every position and for every selection method. In the
objectives above, \(S_t\), the teacher weights on that set, and \(M_t\) are
replaced by the corrected set, weights, and their sum. The resulting target
is not the original teacher distribution conditioned on the corrected set;
Appendix~\ref{app:truncation} formalizes this distinction. This empirical
safeguard is Qwen3-specific; other model families may require different
handling of thinking tokens, or none. The teacher-sampled set therefore follows the exact
sampling law only before correction.
Appendix~\ref{app:gumbel} derives this qualification and distinguishes the
procedure from importance sampling.

\subsection{Token weighting and diagnostics}

For sequence \(i\), let \(C_i\) contain supervised tokens from
\texttt{<think>} through \texttt{</think>}, inclusive, and let \(A_i\)
contain subsequent supervised assistant tokens. Given a per-token objective
\(\ell_t\), when both regions are nonempty the sequence loss is
\[
  \mathcal{L}_i
  =(1-\lambda)\frac{1}{|C_i|}\sum_{t\in C_i}\ell_t
  +\lambda\frac{1}{|A_i|}\sum_{t\in A_i}\ell_t.
\]
A naive token average lets the longer reasoning region dominate the loss.
Our companion study found that balancing the reasoning and answer regions
improved downstream results over this unweighted
objective~\cite{Ghita:2026scaling}. Across our \(\lambda\) ablations,
\(\lambda=0.5\) provided a stable common setting, so we use it throughout
the reasoning-trace experiments. When one region is empty, the available
region receives the full sequence weight. The batch objective averages
equally over sequences. We add neither a hard-label language-modeling loss
nor an intermediate-feature loss.

We report the mean training objective over the final 20 updates. For the
primary ladder, we also evaluate full-vocabulary forward KL after training
on a fixed sample, using all \(V\) classes and the same sequence weighting.
This provides a common diagnostic of disagreement with the complete teacher
distribution; generation metrics determine model quality. Without shuffling
the reasoning-trace training data, we apply the training tokenization and
length filter and retain the first 256 surviving examples (430,657
supervised positions) for every primary-ladder checkpoint. The larger-pair
confirmation uses the analogous diagnostic on eight examples.

\section{Experimental setup}
\label{sec:setup}

\subsection{Data}

Our primary corpus comprises the 545,844 English \texttt{chat} records
marked \texttt{reasoning=on} in the Nemotron Post-Training Dataset
v2~\cite{NemotronPostTraining:2025}. We preserve assistant reasoning traces
and discard 428 records with unclosed think markup. This leaves 545,416
records before tokenization, including 545,160 with nonempty reasoning
spans.

We use the Qwen3 tokenizer and a chat template that preserves assistant
content verbatim. The loss mask includes assistant content and the
end-of-turn token; system content, user content, and assistant role headers
are excluded. Sequences longer than 8,192 tokens are dropped rather than
truncated, removing 9,158 records and leaving 536,258 training examples
with 905,730,102 supervised next-token positions.

The separate answer-only experiment uses the training split of
Puzzle-KD-Nemotron-Post-Training-Dataset-v2~\cite{PuzzleKD:2026}. During
preprocessing, closed reasoning spans are removed from each assistant turn,
and the first bracketed answer is retained when present. Empty outputs and
malformed think spans are discarded. After the 8,192-token length filter,
808,712 training examples remain. The resulting corpus contains
\texttt{chat}, \texttt{code}, \texttt{math}, and \texttt{stem} examples,
whereas the reasoning corpus contains only \texttt{chat} examples. We
therefore treat Puzzle-KD as a separate training setting rather than a
controlled trace-removal ablation.

In both settings, the corpus supplies training contexts and supervised
assistant positions. The frozen Qwen3-4B model supplies the teacher
distribution for the primary ladder and the answer-only experiment. The
larger-pair confirmation uses a frozen Qwen3-8B teacher.

\subsection{Training and evaluation}

We perform full-parameter distillation from a frozen Qwen3-4B teacher to a
trainable Qwen3-1.7B student~\cite{Qwen3:2025}. The models share a vocabulary
of 151,936 tokens. Training uses bfloat16 precision and PyTorch Fully
Sharded Data Parallel v2 (FSDP2) on 64 NVIDIA GB200 GPUs (16 nodes) within a
single NVLink domain.

All primary ladder runs use AdaGrad~\cite{Duchi:2011aa} with a constant
learning rate of \(10^{-5}\), zero weight decay, no warmup, and global
gradient clipping at 1.0. We selected this configuration in a preliminary
sweep over AdaGrad learning rates and AdamW~\cite{Loshchilov:2019adamw}
hyperparameters. The best AdamW configuration favored \(\beta_2=0.999\)
without warmup, consistent with a long second-moment history, but produced
slightly lower MMLU-Pro than AdaGrad. We hold the optimizer fixed across
support sizes.

We train for one epoch (a two-epoch run did not improve MMLU-Pro). The
per-device batch size is 2 and the global batch size is 128, giving 4,189
updates per epoch. All primary runs use \(\tau=1\), \(\lambda=0.5\), and
KL-only distillation, with no hard-label cross-entropy loss.
Rank-renormalized KL is the primary configuration. The partial-mass
objective, teacher sampling, and uniform sampling provide supporting
comparisons. We run the primary rank-\(k\) ladder over three seeds.
Full-vocabulary distillation is the primary reference, while the frozen
Qwen3-1.7B checkpoint provides context for absolute model quality.
The answer-only and additional-model checks follow the same one-epoch,
three-seed protocol with AdaGrad configurations selected for their
respective settings.

We also train Qwen3-4B students from a frozen Qwen3-8B teacher on the same
reasoning corpus, AdaGrad learning rate, \(\lambda=0.5\), and three seeds,
for one epoch at global batch size 144 on 18 nodes (3,724 updates).
Rank-renormalized \(k=15\) and \(k=152\) use per-device batch size 2; the
matched full-vocabulary students use per-device batch size 1 after the
larger microbatch exhausted device memory. Peak allocation and step time
are therefore not compared across those microbatches. The frozen Qwen3-4B
checkpoint is an absolute reference for that pair, not the truncation
control.

As an auxiliary cross-family and cross-architecture check, we distill
NVIDIA-Nemotron-Nano-9B-v2 from frozen NVIDIA-Nemotron-Nano-12B-v2, hybrid
Mamba--Transformer models~\cite{NVIDIA:2025nemotronnano2}. We compare
full-vocabulary training with rank-renormalized support at 0.1\% (\(k=131\)),
0.02\% (\(k=26\)), and 0.01\% (\(k=13\)), evaluating generation with
reasoning both enabled and disabled.

We select MMLU-Pro and GPQA-Diamond as the study concerns reasoning
language models. MMLU-Pro provides broad coverage of knowledge-intensive
questions with a stronger emphasis on reasoning than the original
benchmark~\cite{MMLU-Pro:2024}. It contains 12,032 examples and is our
primary benchmark. GPQA-Diamond complements it with 198 expert-validated
graduate-level science questions~\cite{Rein:2024gpqa} and serves as a
smaller secondary benchmark.

The primary evaluation uses the Qwen3 chat template with thinking
enabled, which we call think-on evaluation. We use greedy decoding, at most
8,192 generated tokens, and a 16,384-token model context. This template
setting is specific to Qwen3 and is distinct from the
\texttt{reasoning=on} metadata used to select training data. Absolute
training time and milliseconds per step are reported; efficiency
comparisons emphasize runs with a shared hardware and software configuration.

\section{Results}
\label{sec:results}

We first evaluate whether retained-support distillation causes observed
degradation in reasoning generation relative to full-vocabulary KL. We then
examine answer-only supervision, additional model pairs, and the systems
consequences of an offline retained-support pipeline.
Training losses and full-vocabulary KL are reported as diagnostics rather
than model-selection criteria.

\subsection{Generation}
\label{sec:generation}

All models in this subsection are evaluated with Qwen3 thinking enabled:
the chat template permits a generated reasoning trace before the final
answer.

We compare each rank-\(k\) student with a full-vocabulary student trained
under the same data, optimization, and update budget.
Table~\ref{tab:k-ladder} reports three-seed means and sample standard
deviations for the final-20 training KL, train time, step time, MMLU-Pro,
and GPQA-Diamond at each \(k\).
Figure~\ref{fig:k-ladder} plots the same generation scores in two panels and
includes the frozen Qwen3-1.7B checkpoint as an absolute reference. The
full-vocabulary student isolates the effect of truncation.

No MMLU-Pro degradation is observed across the ladder: every mean is within
0.0033 of the full-vocabulary score of \(0.474\pm0.003\), and \(k=15\)
obtains \(0.477\pm0.001\). With three seeds, this is not a claim of
statistical equivalence. All distilled students remain below the frozen
Qwen3-1.7B reference on MMLU-Pro. This absolute drop reflects distribution
drift from training on the distillation corpus; because the matched
full-vocabulary student also exhibits it, it is not attributable to
truncation. The experiment tests truncation relative to full-vocabulary
distillation, not improvement over initialization.
GPQA-Diamond is \(0.350\pm0.023\) at full \(V\) and \(0.362\pm0.018\) at
\(k=15\). Given its 198 examples and seed variation, we do not rank nearby
configurations on GPQA-Diamond.

\noindent\begin{minipage}{\linewidth}
\centering
\small
\begin{tabular}{lrrrrrr}
\toprule
\% of \(V\) & \(k\) & Final-20 KL & Train (min) & ms/step & MMLU-Pro & GPQA-Diamond \\
\midrule
100 & 151936 (full) & \(0.424\mathbin{\pm}0.001\) & \(59.5\mathbin{\pm}1.6\) & \(853\mathbin{\pm}23\) & \(0.474\mathbin{\pm}0.003\) & \(0.350\mathbin{\pm}0.023\) \\
50 & 75968 & \(0.424\mathbin{\pm}0.001\) & \(94.8\mathbin{\pm}3.2\) & \(1358\mathbin{\pm}46\) & \(0.477\mathbin{\pm}0.002\) & \(0.345\mathbin{\pm}0.021\) \\
25 & 37984 & \(0.424\mathbin{\pm}0.001\) & \(70.1\mathbin{\pm}2.4\) & \(1005\mathbin{\pm}34\) & \(0.475\mathbin{\pm}0.002\) & \(0.347\mathbin{\pm}0.013\) \\
10 & 15194 & \(0.423\mathbin{\pm}0.001\) & \(62.6\mathbin{\pm}1.1\) & \(897\mathbin{\pm}16\) & \(0.476\mathbin{\pm}0.006\) & \(0.320\mathbin{\pm}0.034\) \\
5 & 7597 & \(0.423\mathbin{\pm}0.001\) & \(62.4\mathbin{\pm}1.5\) & \(894\mathbin{\pm}22\) & \(0.477\mathbin{\pm}0.004\) & \(0.352\mathbin{\pm}0.025\) \\
1 & 1519 & \(0.422\mathbin{\pm}0.001\) & \(58.3\mathbin{\pm}0.8\) & \(835\mathbin{\pm}12\) & \(0.475\mathbin{\pm}0.002\) & \(0.345\mathbin{\pm}0.031\) \\
0.1 & 152 & \(0.412\mathbin{\pm}0.001\) & \(59.9\mathbin{\pm}2.5\) & \(858\mathbin{\pm}35\) & \(0.476\mathbin{\pm}0.003\) & \(0.323\mathbin{\pm}0.053\) \\
0.01 & 15 & \(0.375\mathbin{\pm}0.001\) & \(58.7\mathbin{\pm}1.6\) & \(840\mathbin{\pm}23\) & \(0.477\mathbin{\pm}0.001\) & \(0.362\mathbin{\pm}0.018\) \\
\bottomrule

\end{tabular}
\captionof{table}{\textbf{Rank-\(k\) reasoning ladder.} Three-seed means
\(\pm\) sample standard deviations. Final-20 KL averages the training
objective over the last 20 updates. Times cover one epoch on the shared
16-node GB200 configuration. GPQA-Diamond contains 198 examples and is not
used to rank nearby configurations.}
\label{tab:k-ladder}
\end{minipage}

Tokenization takes 67--72 minutes across the runs and has no systematic
dependence on \(k\); Table~\ref{tab:k-ladder} therefore reports training time
separately. Truncation does not avoid the live teacher's full-vocabulary
computation, and top-\(k\) selection adds overhead. Step time is therefore
nonmonotonic: \(840\pm23\) ms at \(k=15\) versus \(853\pm23\) ms for
full-vocabulary KL. Truncating the objective alone provides no material
speedup.

\noindent\begin{minipage}{\linewidth}
\centering
\includegraphics[width=0.9312\linewidth]{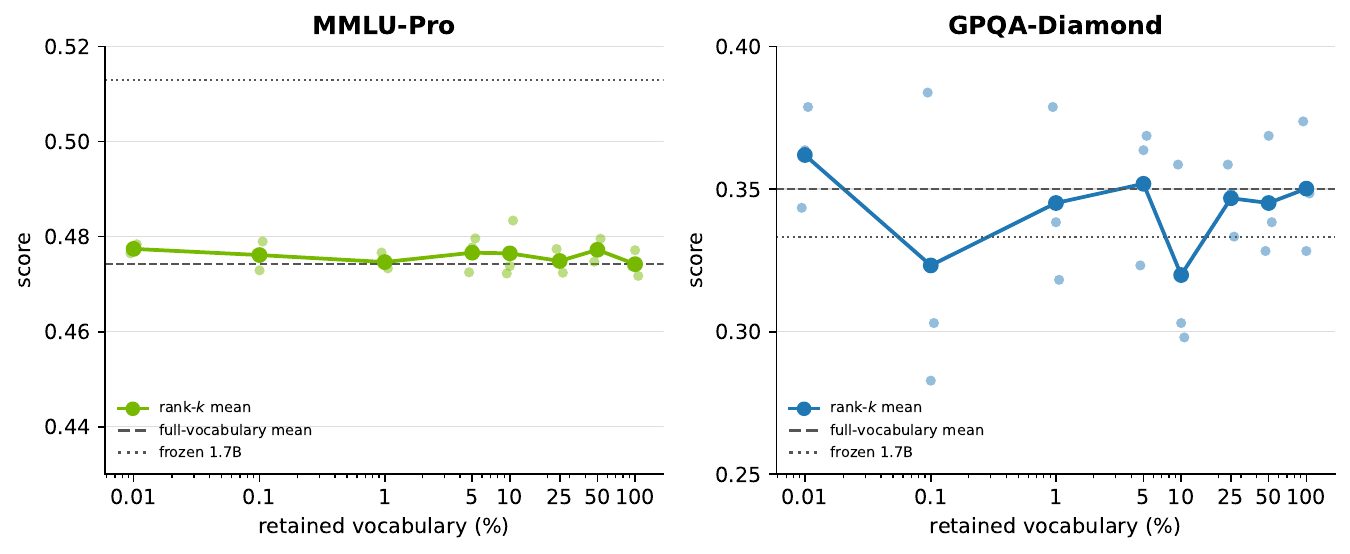}
\captionof{figure}{\textbf{Generation across retained support size.}
Three-seed MMLU-Pro (left) and GPQA-Diamond (right). Lines and large markers
show means; light points show individual seeds. Horizontal lines mark the
full-vocabulary student and frozen Qwen3-1.7B. Both panels use a log-scaled
support axis but different vertical ranges.}
\label{fig:k-ladder}
\end{minipage}

Figure~\ref{fig:train-progress} shows the optimized training objective
throughout the epoch for one fixed seed across all support sizes. The
full-vocabulary through 5\% curves follow nearly the same trajectory, while
the smallest supports converge to lower values of their renormalized
objective. Because the objective changes with \(k\), these vertical
differences show within-objective convergence rather than model quality and
place the final-20 statistic in context.

Figures~\ref{fig:k-ladder}--\ref{fig:kl-diagnostics} distinguish
the optimized conditional objective, agreement with the full teacher
distribution, and generation quality. Renormalized KL matches only relative
probabilities within the retained set. Shrinking \(k\) removes constraints
on the tail and on the total probability assigned to that set, so a lower
training objective can coincide with greater disagreement with the full
teacher distribution. Generation asks a different question. Under greedy
decoding, it depends on whether the highest-probability choices along the
generated trajectory change, not on whether the entire softmax is calibrated.
For the concentrated teacher distributions studied here, this provides
intuition for stable benchmark performance through \(k=15\) despite that
increased disagreement. Training loss and full-vocabulary KL are therefore
useful diagnostics, but neither substitutes for generation.

\noindent\begin{minipage}{\linewidth}
\centering
\includegraphics[width=0.7372\linewidth]{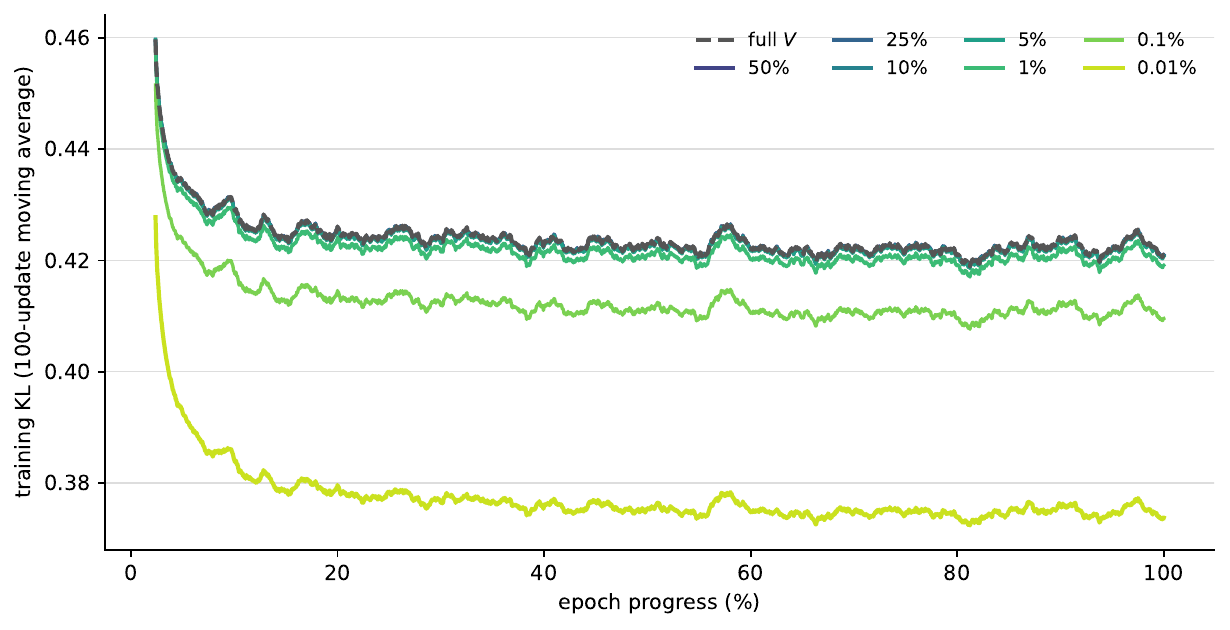}
\captionof{figure}{\textbf{Training KL throughout one epoch.} Curves are
100-update moving averages of the optimized objective for one fixed seed,
used across every \(k\). Table~\ref{tab:k-ladder} and
Figure~\ref{fig:k-ladder} aggregate three seeds. Each \(k\) defines a
different renormalized objective, so vertical values are not a common
measure of model quality.}
\label{fig:train-progress}
\end{minipage}

Table~\ref{tab:k-ladder} and Figure~\ref{fig:train-progress} report the
\(k\)-dependent optimized objective. To compare checkpoints under a common
objective, Figure~\ref{fig:kl-diagnostics} evaluates an in-sample
full-vocabulary KL diagnostic; Appendix~\ref{app:truncation} identifies the
terms retained and omitted by the truncated objective.

\noindent\begin{minipage}{\linewidth}
\centering
\includegraphics[width=0.7372\linewidth]{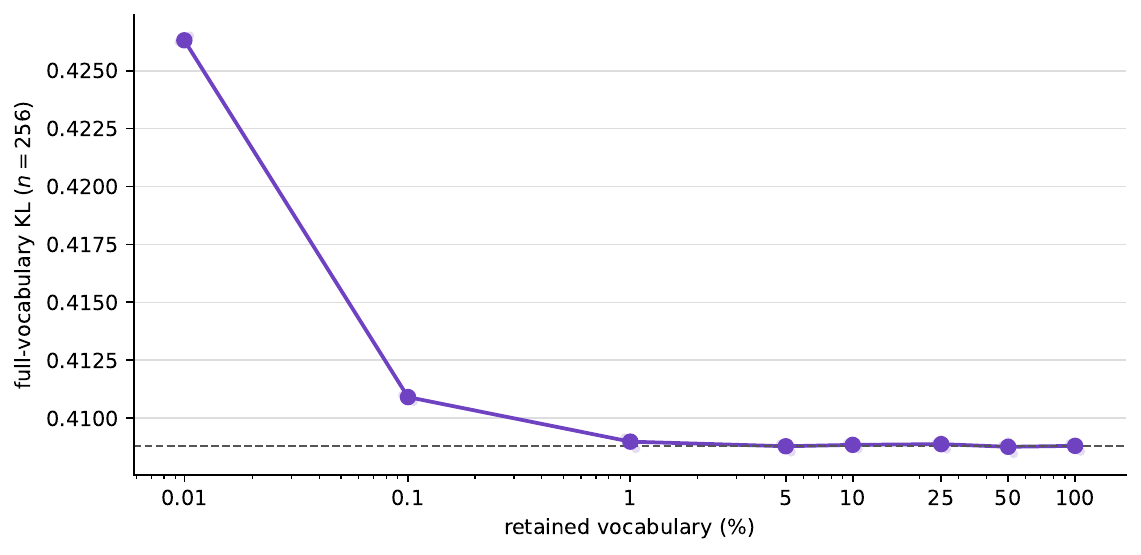}
\captionof{figure}{\textbf{Full-vocabulary KL across retained support size.}
Three-seed in-sample diagnostic on the same 256 training examples (430,657
supervised positions). The line shows the mean, light points show individual
seeds, and the dashed line marks the full-\(V\) mean.}
\label{fig:kl-diagnostics}
\end{minipage}

\subsection{Answer-only supervision}
\label{sec:answer-only}

To test whether very small support remains effective without reasoning
traces in the training targets, we repeat the rank-renormalized study on the
answer-only Puzzle-KD corpus~\cite{PuzzleKD:2026}. It uses the same teacher
and student but different source examples and categories.

We train a full-vocabulary reference and rank-\(k\) students at 50\%, 10\%,
1\%, and 0.01\% of \(V\), over three seeds. The training loss is the mean
rank-renormalized KL over supervised answer tokens. Puzzle-KD contains no
reasoning region, so all sequence weight belongs on the answer region
(\(\lambda=1\)); there is no second region to balance. At evaluation, Qwen3
thinking disabled produces a direct answer, while thinking enabled permits a
model-generated reasoning trace before the answer. These are Qwen3-specific
template settings; reporting both separates the support-size result from the
choice of generation mode.

\begin{center}
\centering
\includegraphics[width=0.9312\linewidth]{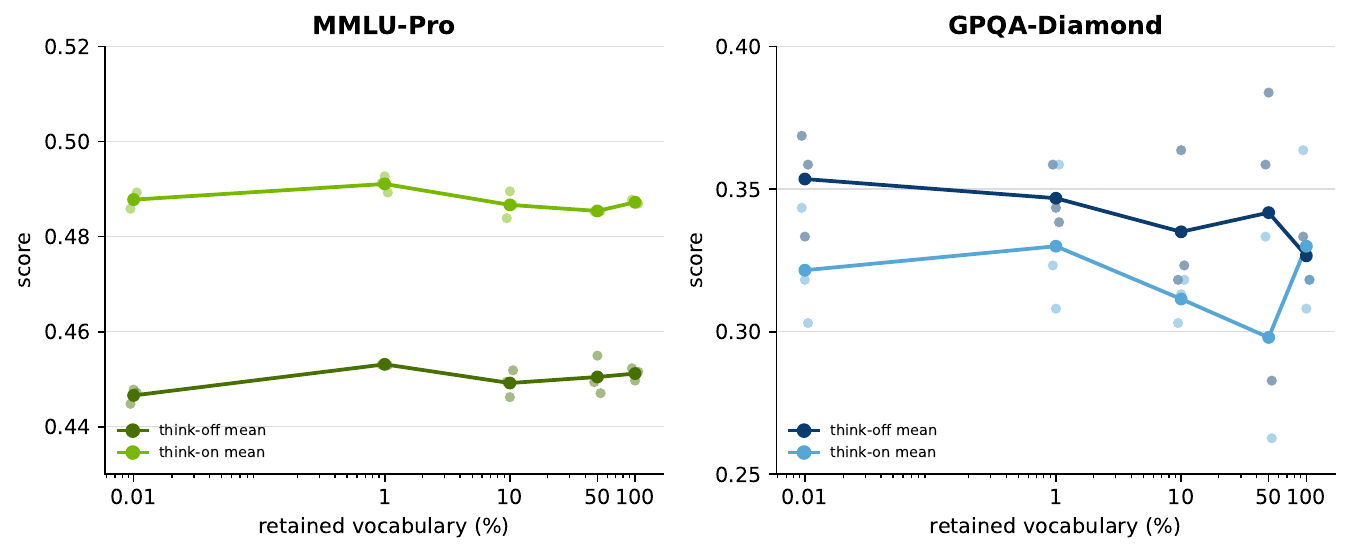}
\captionof{figure}{\textbf{Generation with answer-only supervision.}
Three-seed MMLU-Pro (left) and GPQA-Diamond (right) with Qwen3 thinking
disabled and enabled. Lines and large markers show means; light points show
individual seeds. The rightmost point is full-vocabulary distillation.}
\label{fig:without-traces}
\end{center}

The support-size result from Section~\ref{sec:generation} carries over to answer-only
supervision: in both evaluation modes, every MMLU-Pro mean remains within
0.005 of its full-vocabulary reference through \(k=15\). The separation
between thinking enabled and disabled is much larger than the variation with
\(k\), but the support-size conclusion is the same within each mode. Because
Puzzle-KD supplies no reasoning traces, the generated reasoning in the
enabled mode is not directly supervised by this corpus. GPQA-Diamond varies
substantially across seeds and shows no consistent support-size trend, so we
report it as a secondary result without ranking configurations.

\subsection{Additional model pairs}

Two additional teacher--student pairs test whether the retained-support
result extends beyond the primary Qwen3 pair. For Qwen3-4B distilled from
frozen Qwen3-8B, retaining 0.1\% (\(k=152\)) or 0.01\% (\(k=15\)) of the
vocabulary produces no observed MMLU-Pro degradation relative to
full-vocabulary training across three seeds; at the smaller support,
the eight-example full-vocabulary KL diagnostic worsens even as the optimized
objective falls, reinforcing the distinction between matching the teacher's
tail and preserving generation. For the auxiliary Nemotron pair, retaining 0.1\%
(\(k=131\)) keeps MMLU-Pro within 0.004 of full-vocabulary training in both
reasoning modes, whereas both 0.01\% (\(k=13\)) and 0.02\% (\(k=26\)) fail
specifically to emit the closing reasoning tag, preventing valid benchmark
evaluation and showing that the smallest reliable support is model-family
dependent.

\subsection{Offline targets and compact output computation}
\label{sec:offline}

The \(k=15\) ladders in Sections~\ref{sec:generation}
and~\ref{sec:answer-only} show no observed MMLU-Pro degradation while
retaining only 0.01\% of the vocabulary. This small support makes the teacher
practical to cache. Before student training, we run the frozen Qwen3-4B
teacher once over the tokenized corpus. At each supervised position, we
retain the top-\(k\) token identifiers and corresponding teacher logits,
together with the logits for both Qwen3 reasoning tags, and write them to
disk. The resulting cache for all 536,258 reasoning examples requires
approximately 108 GiB (116 GB); student training reads these targets instead
of executing the teacher, removing the teacher forward pass from the training
loop. With these targets cached, the renormalized objective also permits
student output computation on only the cached rows. Sparse offline
targets follow prior work~\cite{Shum:2024first,Anshumann:2025sls,
HaoPeng:2025pretraining,Ryskulov:2026offline}.

Table~\ref{tab:offline-training} shows that, across three matched seeds,
caching reduces mean step time by 38\%. Compact output reduces peak allocation
by 24\% relative to cached full-head training and by 46\% relative to
live-teacher training, but does not reduce step time further. All three
final-20 objectives remain close. Tokenization remains 69--70 minutes.

\noindent\begin{minipage}{\linewidth}
\centering
\small
\begin{tabular}{lcccc}
\toprule
Configuration & Final-20 loss & Train (min) & ms/step & Peak alloc. (GB) \\
\midrule
Live teacher + full output head & \(0.3754\mathbin{\pm}0.0013\) & \(58.7\mathbin{\pm}1.6\) & \(840\mathbin{\pm}23\) & \(103.3\mathbin{\pm}0.2\) \\
Cached teacher + full output head & \(0.3754\mathbin{\pm}0.0013\) & \(36.2\mathbin{\pm}1.8\) & \(519\mathbin{\pm}26\) & \(73.4\mathbin{\pm}0.2\) \\
Cached teacher + compact \(k\)-row head & \(0.3773\mathbin{\pm}0.0013\) & \(36.1\mathbin{\pm}2.5\) & \(516\mathbin{\pm}35\) & \(55.8\mathbin{\pm}0.3\) \\
\bottomrule

\end{tabular}
\captionof{table}{\textbf{Matched full-epoch \(k=15\) training.} Means
\(\pm\) sample standard deviations over three seeds. Peak allocation is the
maximum post-warmup CUDA allocation across GPUs.}
\label{tab:offline-training}
\end{minipage}

On MMLU-Pro, live-teacher, cached full-head, and compact \(k\)-row training
obtain \(0.477\pm0.001\), \(0.478\pm0.003\), and \(0.476\pm0.002\),
respectively. Both cached means are within 0.0014 of live-teacher training,
so no MMLU-Pro degradation is observed. The corresponding GPQA-Diamond
scores are \(0.362\pm0.018\), \(0.337\pm0.042\), and
\(0.328\pm0.022\). Given the benchmark's 198 examples and seed variation,
we report these values without ranking configurations.

Cached full-head training still materializes student logits of shape
\((B,S,V)\). Compact training instead projects the \(N\) valid supervised
positions onto only the cached rows, producing
\((N,\widetilde{k})\) logits with \(\widetilde{k}\leq k+2\) after the Qwen3
support correction. Appendix~\ref{app:truncation} shows why this restriction
is valid for renormalized KL but not for the partial-mass objective. Without
a cache, the teacher must still score every vocabulary row to identify the
retained set, so restricting only the student output does not remove the
teacher-side output pass. After caching, the full student head uses one
highly optimized dense GEMM, whereas compact output uses indexed weight-row
gathers and batched narrow products. Their data movement and lower kernel
efficiency offset the reduction in arithmetic, leaving step time unchanged
while reducing peak allocation. Thus caching and compact output computation
address wall time and memory, respectively, rather than compounding into the
same speedup.

\section{Conclusion}
\label{sec:conclusion}

The native Qwen3-4B
distribution is highly concentrated, and the three-seed rank-\(k\) ladder
shows no observed MMLU-Pro degradation relative to full-vocabulary
distillation down to \(k=15\), or 0.01\% of the vocabulary. Distilling
Qwen3-4B from Qwen3-8B reproduces the MMLU-Pro result at \(k=15\) and
\(k=152\). The Nemotron check retains MMLU-Pro at 0.1\% of the vocabulary
but exposes a reasoning-termination failure at smaller supports, so the
cutoff itself does not transfer universally across model families.
Answer-only Puzzle-KD shows the same support-size trend under both Qwen3
evaluation modes, although it is a separate training setting rather than a
matched ablation. Full-vocabulary KL nevertheless increases at the
smallest supports, distinguishing the support needed to match the complete
teacher distribution from that needed for benchmark generation.

This small effective support enables a practical offline pipeline. Caching
the selected \(k=15\) teacher identifiers and logits removes the teacher
from training and reduces step time by 38\%. Renormalized KL also permits
compact \(k\)-row student output computation, which avoids materializing
vocabulary-sized logits and reduces peak allocated memory by 24\% relative
to cached full-head training, although it provides no additional speedup.
Both cached variants show no observed MMLU-Pro degradation relative to
live-teacher training. Caching and compact output computation therefore
address distinct bottlenecks: training time and memory, respectively.

The primary conclusions are scoped to two Qwen3 teacher--student pairs
(1.7B from 4B; 4B from 8B), with one auxiliary Nemotron family check, one
training epoch, and the two corpora studied here. We do not locate the Qwen3
failure boundary below \(k=15\). The structural-token correction and
thinking-enabled evaluation are specific to Qwen3, while Puzzle-KD provides
a separate answer-only setting rather than a controlled trace-removal
ablation. The small GPQA-Diamond benchmark and in-sample full-vocabulary KL
diagnostic provide supporting evidence, and the systems measurements depend
on the hardware and implementation. More broadly, effective support should
be measured for the teacher, data, and temperature of interest, then
validated through generation. Once validated, it provides a direct path
from concentrated teacher predictions to smaller offline targets and lower
distillation cost.

\noindent\textbf{Acknowledgments.}\enspace
We thank Saleh Ashkboos and Liana Mikaelyan for helpful discussions and
reviews.
\par

\clearpage
\section*{Appendix}
\setcounter{subsection}{0}
\renewcommand{\thesubsection}{\Alph{subsection}}

\subsection{What retained-support objectives optimize}
\label{app:truncation}

We first decompose full-vocabulary KL at a fixed temperature, suppressing the
common \(\tau^2\) factor, for a fixed, unmodified support \(S\). Let
\[
  M=\sum_{v\in S}p_v,
  \qquad
  Q=\sum_{v\in S}q_v,
\]
and define the conditional distributions
\[
  p_S(v)=\frac{p_v}{M},
  \quad
  q_S(v)=\frac{q_v}{Q}
  \quad (v\in S),
\]
with \(p_{\bar S}\) and \(q_{\bar S}\) defined analogously on the complement.
For a nonempty proper subset \(S\), softmax probabilities give
\(0<M,Q<1\). Splitting the KL sum over \(S\) and \(\bar S\) yields
\[
\begin{aligned}
  \sum_{v\in S}p_v\log\frac{p_v}{q_v}
  &=
  M\,\operatorname{KL}(p_S\,\|\,q_S)
  +M\log\frac{M}{Q},\\
  \sum_{v\notin S}p_v\log\frac{p_v}{q_v}
  &=
  (1-M)\,\operatorname{KL}(p_{\bar S}\,\|\,q_{\bar S})
  +(1-M)\log\frac{1-M}{1-Q}.
\end{aligned}
\]
The Bernoulli KL between the teacher and student masses assigned to \(S\)
is
\[
  \operatorname{KL}\!\left(
    \operatorname{Bern}(M)\,\|\,\operatorname{Bern}(Q)
  \right)
  =
  M\log\frac{M}{Q}
  +(1-M)\log\frac{1-M}{1-Q}.
\]
Therefore,
\[
\begin{aligned}
  \operatorname{KL}(p\,\|\,q)
  ={}&
  M\,\operatorname{KL}(p_S\,\|\,q_S)
  +(1-M)\,\operatorname{KL}(p_{\bar S}\,\|\,q_{\bar S})\\
  &+
  \operatorname{KL}\!\left(
    \operatorname{Bern}(M)\,\|\,\operatorname{Bern}(Q)
  \right).
\end{aligned}
\]
The identity extends to \(S=V\) by continuity. Renormalized KL optimizes
only \(\operatorname{KL}(p_S\|q_S)\). It discards both the conditional
tail term and the binary term that matches total mass on \(S\).
The partial-mass objective is exactly
\[
  \mathcal{L}_{\mathrm{mass}}
  =
  M\,\operatorname{KL}(p_S\,\|\,q_S)
  +M\log\frac{M}{Q}.
\]
It retains pressure on the student's total selected mass \(Q\), but omits
the conditional tail and the complementary contribution
\((1-M)\log((1-M)/(1-Q))\).

\paragraph{Student-logit gradients.}
Let \(q^\tau=\operatorname{softmax}(z^{\mathrm S}/\tau)\), and treat the
teacher distribution and selected set as fixed. Since
\[
  \frac{\partial\log q_v^\tau}{\partial z_j^{\mathrm S}}
  =
  \frac{1}{\tau}
  \left(\mathbf{1}\{j=v\}-q_j^\tau\right),
\]
the \(\tau^2\)-scaled full-vocabulary objective has gradient
\[
  \frac{\partial\mathcal{L}_{\mathrm{full}}}
       {\partial z_j^{\mathrm S}}
  =
  \tau(q_j^\tau-p_j^\tau).
\]
For renormalized KL,
\[
  \frac{\partial\mathcal{L}_{\mathrm{renorm}}}
       {\partial z_j^{\mathrm S}}
  =
  \begin{cases}
    \tau(\bar q_j-\bar p_j), & j\in S,\\
    0, & j\notin S.
  \end{cases}
\]
For the partial-mass objective,
\[
  \frac{\partial\mathcal{L}_{\mathrm{mass}}}
       {\partial z_j^{\mathrm S}}
  =
  \tau\left(
    Mq_j^\tau-p_j^\tau\mathbf{1}\{j\in S\}
  \right),
\]
as also derived for cached partial-mass targets by
Ryskulov et al.~\cite{Ryskulov:2026offline}. Thus renormalized KL has no
student-logit gradient outside \(S\), whereas the partial-mass objective
still acts on all \(V\) logits through the full-vocabulary normalizer.

\paragraph{Why rank selection maximizes retained mass.}
Let \(\pi\) be a permutation of the token identifiers such that
\[
  p_{\pi(1)}\geq p_{\pi(2)}\geq\cdots\geq p_{\pi(V)},
\]
and define \(S_k^{\mathrm{rank}}=\{\pi(1),\ldots,\pi(k)\}\). Consider any
set \(S\subseteq\{1,\ldots,V\}\) with \(|S|=k\). If
\(S\neq S_k^{\mathrm{rank}}\), equal cardinality implies that some
\(\pi(i)\) with \(i\leq k\) is absent from \(S\), while some \(\pi(j)\)
with \(j>k\) is present. Replacing \(\pi(j)\) by \(\pi(i)\) changes the
retained mass by
\[
  p_{\pi(i)}-p_{\pi(j)}\geq0.
\]
Repeating this exchange finitely many times transforms \(S\) into
\(S_k^{\mathrm{rank}}\) without decreasing its mass. Therefore,
\[
  S_k^{\mathrm{rank}}
  \in
  \operatorname*{arg\,max}_{\substack{S\subseteq\{1,\ldots,V\}\\|S|=k}}
  \sum_{v\in S}p_v,
\]
and another size-\(k\) set can attain the same mass only when teacher
probabilities are tied across the selection boundary.

\paragraph{Qualification for the Qwen3 correction.}
The decomposition above is exact for a literal restriction of \(p\) to
\(S\). The Qwen3 correction enlarges the set and modifies inserted teacher
weights. The gradient forms remain valid after replacing \(S\), the teacher
weights, and their total mass by their corrected counterparts, but the
result is no longer a decomposition of full KL for the original teacher
distribution.

\subsection{Gumbel-top-\texorpdfstring{\(k\)}{k} support sampling}
\label{app:gumbel}

This appendix derives the sampling law used by the teacher-sampled support
check. For one supervised position, define
\[
  a_v=z_v^{\mathrm T}/\tau,
  \qquad
  w_v=\exp(a_v),
  \qquad
  p_v=\frac{w_v}{\sum_u w_u}.
\]
Let \(E_v\sim\operatorname{Exp}(1)\) independently and set
\(g_v=-\log E_v\). Since
\[
  \Pr(g_v\leq x)
  =\Pr(E_v\geq e^{-x})
  =\exp(-e^{-x}),
\]
each \(g_v\) is a standard Gumbel random variable. The perturbed score can be
written as
\[
  a_v+g_v
  =\log w_v-\log E_v
  =-\log\left(\frac{E_v}{w_v}\right).
\]
The random variable \(T_v=E_v/w_v\) is exponential with rate \(w_v\).
Sorting the perturbed scores from largest to smallest is therefore
equivalent to sorting the independent arrival times \(T_v\) from smallest
to largest.

The probability that token \(i\) arrives first is
\[
  \Pr\!\left(T_i=\min_v T_v\right)
  =\int_0^\infty
    w_i e^{-w_i t}\prod_{v\ne i}e^{-w_v t}\,dt
  =\frac{w_i}{\sum_v w_v}
  =p_i.
\]
After an arrival, the memoryless property of exponential random variables
gives the same race over the remaining tokens. Hence the probability of an
ordered sample \((i_1,\ldots,i_k)\) is
\[
  \Pr(i_1,\ldots,i_k)
  =
  \prod_{j=1}^{k}
  \frac{w_{i_j}}
       {\sum_{v\notin\{i_1,\ldots,i_{j-1}\}}w_v}.
\]
This is the Plackett--Luce distribution over rankings
\cite{Luce:1959choice,Plackett:1975permutations}. Let \(\pi^g\) order the
perturbed scores \(z_v^{\mathrm T}/\tau+g_v\) from largest to smallest.
Then, up to events of probability zero from ties,
\((\pi^g(1),\ldots,\pi^g(k))\) is an exact ordered sample of \(k\) distinct
tokens without replacement from the teacher distribution, with unordered
support \(S=\{\pi^g(1),\ldots,\pi^g(k)\}\)~\cite{Kool:2019gumbel}.

\paragraph{Qwen3 support correction.}
The exact sampling statement applies to the raw set \(S\). Our loss then
adds each Qwen3 think-tag identifier when it is absent
\cite{Qwen3:2025}, producing a corrected set
\(\widetilde S\) with \(k\), \(k+1\), or \(k+2\) elements. It also modifies
the inserted teacher weights as described in Section~\ref{sec:method}.
Consequently, \(\widetilde S\) is not itself distributed according to the
Gumbel-top-\(k\) law.

\paragraph{Relation to importance sampling.}
Even before the support correction, sampling \(S\) does not make the
resulting truncated loss an unbiased estimator of full-vocabulary KL. For
the partial-mass objective, let
\[
  h_v=p_v\log\frac{p_v}{q_v},
  \qquad
  \iota_v=\Pr(v\in S).
\]
Then
\[
  \mathbb{E}_S\left[\sum_{v\in S}h_v\right]
  =\sum_{v=1}^{V}\iota_v h_v,
\]
which differs from \(\sum_v h_v\) unless the terms are corrected by their
inclusion probabilities. The renormalized objective is further nonlinear
in \(S\), because both teacher and student normalizers depend on the sampled
set:
\[
  \mathbb{E}_S\!\left[
    \operatorname{KL}\bigl(p(\cdot\mid S)\,\|\,q(\cdot\mid S)\bigr)
  \right]
  \ne
  \operatorname{KL}(p\,\|\,q)
\]
in general. Random Sampling KD instead uses proposal corrections to preserve
the full-KL gradient in expectation~\cite{Anshumann:2025sls}. Our
teacher-sampled variant uses no such correction and is only a robustness
check on support construction.

\phantomsection
\addcontentsline{toc}{section}{References}
\begingroup
\footnotesize
\urlstyle{same}
\setlength{\emergencystretch}{3em}
\bibliographystyle{unsrtnat}
\bibliography{refs}

\begin{thebibliography}{27}
\providecommand{\natexlab}[1]{#1}
\providecommand{\url}[1]{\texttt{#1}}
\expandafter\ifx\csname urlstyle\endcsname\relax
  \providecommand{\doi}[1]{doi: #1}\else
  \providecommand{\doi}{doi: \begingroup \urlstyle{rm}\Url}\fi

\bibitem[Hinton et~al.(2015)Hinton, Vinyals, and Dean]{Hinton:2015aa}
Geoffrey Hinton, Oriol Vinyals, and Jeff Dean.
\newblock Distilling the knowledge in a neural network.
\newblock 2015.
\newblock URL \url{https://arxiv.org/abs/1503.02531}.

\bibitem[Gu et~al.(2024)Gu, Dong, Wei, and Huang]{Gu:2023aa}
Yuxian Gu, Li~Dong, Furu Wei, and Minlie Huang.
\newblock {MiniLLM}: Knowledge distillation of large language models.
\newblock In \emph{International Conference on Learning Representations}, 2024.
\newblock URL
  \url{https://proceedings.iclr.cc/paper_files/paper/2024/hash/8ac015d409635f196f9e3e9dcfb9a94e-Abstract-Conference.html}.

\bibitem[Agarwal et~al.(2024)Agarwal, Vieillard, Zhou, Stanczyk, Ramos, Geist,
  and Bachem]{Agarwal:2024gkd}
Rishabh Agarwal, Nino Vieillard, Yongchao Zhou, Piotr Stanczyk, Sabela Ramos,
  Matthieu Geist, and Olivier Bachem.
\newblock On-policy distillation of language models: Learning from
  self-generated mistakes.
\newblock In \emph{International Conference on Learning Representations
  (ICLR)}, 2024.
\newblock URL \url{https://arxiv.org/abs/2306.13649}.

\bibitem[Ko et~al.(2024)Ko, Kim, Chen, and Yun]{Ko:2024aa}
Jongwoo Ko, Sungnyun Kim, Tianyi Chen, and Se-Young Yun.
\newblock {DistiLLM}: Towards streamlined distillation for large language
  models.
\newblock In \emph{Proceedings of the 41st International Conference on Machine
  Learning}, volume 235 of \emph{Proceedings of Machine Learning Research},
  pages 24872--24895, 2024.
\newblock URL \url{https://proceedings.mlr.press/v235/ko24c.html}.

\bibitem[Shum et~al.(2024)Shum, Xu, Zhang, Chen, Diao, Dong, Zhang, and
  Raza]{Shum:2024first}
KaShun Shum, Minrui Xu, Jianshu Zhang, Zixin Chen, Shizhe Diao, Hanze Dong,
  Jipeng Zhang, and Muhammad~Omer Raza.
\newblock {FIRST}: Teach a reliable large language model through efficient
  trustworthy distillation.
\newblock In \emph{Proceedings of the 2024 Conference on Empirical Methods in
  Natural Language Processing}, pages 12646--12659, 2024.
\newblock \doi{10.18653/v1/2024.emnlp-main.703}.
\newblock URL \url{https://aclanthology.org/2024.emnlp-main.703/}.

\bibitem[Anshumann et~al.(2025)Anshumann, Zaidi, Kedia, Ahn, Kwon, Lee, Lee,
  and Lee]{Anshumann:2025sls}
Anshumann, Mohd~Abbas Zaidi, Akhil Kedia, Jinwoo Ahn, Taehwak Kwon, Kangwook
  Lee, Haejun Lee, and Joohyung Lee.
\newblock Sparse logit sampling: Accelerating knowledge distillation in {LLM}s.
\newblock In \emph{Proceedings of the 63rd Annual Meeting of the Association
  for Computational Linguistics}, pages 18085--18108, 2025.
\newblock \doi{10.18653/v1/2025.acl-long.885}.
\newblock URL \url{https://aclanthology.org/2025.acl-long.885/}.

\bibitem[Wei et~al.(2022)Wei, Wang, Schuurmans, Bosma, Ichter, Xia, Chi, Le,
  and Zhou]{Wei:2022aa}
Jason Wei, Xuezhi Wang, Dale Schuurmans, Maarten Bosma, Brian Ichter, Fei Xia,
  Ed~Chi, Quoc Le, and Denny Zhou.
\newblock Chain-of-thought prompting elicits reasoning in large language
  models.
\newblock In \emph{Advances in Neural Information Processing Systems}, 2022.
\newblock URL
  \url{https://proceedings.neurips.cc/paper/2022/hash/9d5609613524ecf4f15af0f7b31abca4-Abstract-Conference.html}.

\bibitem[Wang et~al.(2024)Wang, Ma, Zhang, Ni, Chandra, Guo, Ren, Arulraj, He,
  Jiang, Li, Ku, Wang, Zhuang, Fan, Yue, and Chen]{MMLU-Pro:2024}
Yubo Wang, Xueguang Ma, Ge~Zhang, Yuansheng Ni, Abhranil Chandra, Shiguang Guo,
  Weiming Ren, Aaran Arulraj, Xuan He, Ziyan Jiang, Tianle Li, Max Ku, Kai
  Wang, Alex Zhuang, Rongqi Fan, Xiang Yue, and Wenhu Chen.
\newblock {MMLU-Pro}: A more robust and challenging multi-task language
  understanding benchmark, 2024.
\newblock URL \url{https://arxiv.org/abs/2406.01574}.

\bibitem[Rein et~al.(2024)Rein, Hou, Stickland, Petty, Pang, Dirani, Michael,
  and Bowman]{Rein:2024gpqa}
David Rein, Betty~Li Hou, Asa~Cooper Stickland, Jackson Petty, Richard~Yuanzhe
  Pang, Julien Dirani, Julian Michael, and Samuel~R. Bowman.
\newblock {GPQA}: A graduate-level {Google}-proof {Q\&A} benchmark.
\newblock \emph{Conference on Language Modeling (COLM)}, 2024.
\newblock URL \url{https://arxiv.org/abs/2311.12022}.

\bibitem[{Qwen Team}(2025)]{Qwen3:2025}
{Qwen Team}.
\newblock {Qwen3} technical report, 2025.
\newblock URL \url{https://arxiv.org/abs/2505.09388}.

\bibitem[{NVIDIA}(2025{\natexlab{a}})]{NVIDIA:2025nemotronnano2}
{NVIDIA}.
\newblock {NVIDIA Nemotron Nano 2}: An accurate and efficient hybrid
  {Mamba--Transformer} reasoning model, 2025{\natexlab{a}}.
\newblock URL \url{https://arxiv.org/abs/2508.14444}.

\bibitem[Muralidharan et~al.(2024)Muralidharan, Sreenivas, Joshi, Chochowski,
  Patwary, Shoeybi, Catanzaro, Kautz, and Molchanov]{Muralidharan:2024minitron}
Saurav Muralidharan, Sharath~Turuvekere Sreenivas, Raviraj Joshi, Marcin
  Chochowski, Mostofa Patwary, Mohammad Shoeybi, Bryan Catanzaro, Jan Kautz,
  and Pavlo Molchanov.
\newblock Compact language models via pruning and knowledge distillation.
\newblock In \emph{Advances in Neural Information Processing Systems}, 2024.
\newblock URL \url{https://arxiv.org/abs/2407.14679}.

\bibitem[Ho et~al.(2023)Ho, Schmid, and Yun]{Ho:2023reasoning}
Namgyu Ho, Laura Schmid, and Se-Young Yun.
\newblock Large language models are reasoning teachers.
\newblock In \emph{Proceedings of the 61st Annual Meeting of the Association
  for Computational Linguistics}, pages 14852--14882, 2023.
\newblock \doi{10.18653/v1/2023.acl-long.830}.
\newblock URL \url{https://aclanthology.org/2023.acl-long.830/}.

\bibitem[Chen et~al.(2026)Chen, Kothapalli, Fatahibaarzi, Sang, Tang, Song,
  Wang, and Abdul-Mageed]{Chen:2026essence}
Wei-Rui Chen, Vignesh Kothapalli, Ata Fatahibaarzi, Hejian Sang, Shao Tang,
  Qingquan Song, Zhipeng Wang, and Muhammad Abdul-Mageed.
\newblock Distilling the essence: Efficient reasoning distillation via sequence
  truncation.
\newblock In \emph{Findings of the Association for Computational Linguistics:
  ACL 2026}, pages 12092--12122, 2026.
\newblock \doi{10.18653/v1/2026.findings-acl.587}.
\newblock URL \url{https://aclanthology.org/2026.findings-acl.587/}.

\bibitem[Raman et~al.(2023)Raman, Mani, Liang, and Lipton]{Raman:2023slim}
Mrigank Raman, Pranav Mani, Davis Liang, and Zachary~C. Lipton.
\newblock For distillation, tokens are not all you need.
\newblock In \emph{NeurIPS 2023 Workshop on Instruction Tuning and Instruction
  Following}, 2023.
\newblock URL \url{https://openreview.net/forum?id=2fc5GOPYip}.

\bibitem[Peng et~al.(2025)Peng, Lv, Bai, Yao, Zhang, Hou, and
  Li]{HaoPeng:2025pretraining}
Hao Peng, Xin Lv, Yushi Bai, Zijun Yao, Jiajie Zhang, Lei Hou, and Juanzi Li.
\newblock Pre-training distillation for large language models: A design space
  exploration.
\newblock In \emph{Proceedings of the 63rd Annual Meeting of the Association
  for Computational Linguistics}, pages 3603--3618, 2025.
\newblock \doi{10.18653/v1/2025.acl-long.181}.
\newblock URL \url{https://aclanthology.org/2025.acl-long.181/}.

\bibitem[Peng and Zhang(2025)]{Peng:2025rlkd}
Tianyu Peng and Jiajun Zhang.
\newblock Enhancing knowledge distillation of large language models through
  efficient multi-modal distribution alignment.
\newblock In \emph{Proceedings of the 31st International Conference on
  Computational Linguistics}, pages 2478--2496, 2025.
\newblock URL \url{https://aclanthology.org/2025.coling-main.169/}.

\bibitem[Tavor et~al.(2026)Tavor, Ebenspanger, Cnaan, and
  Geva]{Tavor:2026selective}
Almog Tavor, Itay Ebenspanger, Neil Cnaan, and Mor Geva.
\newblock Rethinking selective knowledge distillation.
\newblock 2026.
\newblock URL \url{https://arxiv.org/abs/2602.01395}.

\bibitem[Ryskulov et~al.(2026)Ryskulov, Garc{\'i}a-Ferrero, Montero, Jansen,
  Hashemi, Garcia, Tiene, and Or{\'u}s]{Ryskulov:2026offline}
Bakbergen Ryskulov, Iker Garc{\'i}a-Ferrero, David Montero, David Jansen, Ali
  Hashemi, Jezabel~R. Garcia, Antonio Tiene, and Rom{\'a}n Or{\'u}s.
\newblock Efficient knowledge distillation for {LLM}s: Offline top-k logits and
  a fused chunked {KL} loss.
\newblock 2026.
\newblock URL \url{https://arxiv.org/abs/2608.03796}.

\bibitem[Kool et~al.(2019)Kool, van Hoof, and Welling]{Kool:2019gumbel}
Wouter Kool, Herke van Hoof, and Max Welling.
\newblock Stochastic beams and where to find them: The {G}umbel-top-k trick for
  sampling sequences without replacement.
\newblock In \emph{Proceedings of the 36th International Conference on Machine
  Learning}, volume~97, pages 3499--3508, 2019.
\newblock URL \url{https://proceedings.mlr.press/v97/kool19a.html}.

\bibitem[Ghita et~al.(2026)Ghita, Desai, and Boier]{Ghita:2026scaling}
Lavinia Ghita, Dhruv Desai, and Ioana Boier.
\newblock Scaling laws for task-specific {LLM} distillation.
\newblock 2026.
\newblock URL \url{https://arxiv.org/abs/2606.24747}.

\bibitem[{NVIDIA}(2025{\natexlab{b}})]{NemotronPostTraining:2025}
{NVIDIA}.
\newblock {Nemotron} post-training dataset v2.
\newblock
  \url{https://huggingface.co/datasets/nvidia/Nemotron-Post-Training-Dataset-v2},
  2025{\natexlab{b}}.

\bibitem[{NVIDIA}(2026)]{PuzzleKD:2026}
{NVIDIA}.
\newblock {Puzzle-KD-Nemotron-Post-Training-Dataset-v2}.
\newblock
  \url{https://huggingface.co/datasets/nvidia/Puzzle-KD-Nemotron-Post-Training-Dataset-v2},
  2026.

\bibitem[Duchi et~al.(2011)Duchi, Hazan, and Singer]{Duchi:2011aa}
John Duchi, Elad Hazan, and Yoram Singer.
\newblock Adaptive subgradient methods for online learning and stochastic
  optimization.
\newblock \emph{Journal of Machine Learning Research}, 12:\penalty0 2121--2159,
  2011.

\bibitem[Loshchilov and Hutter(2019)]{Loshchilov:2019adamw}
Ilya Loshchilov and Frank Hutter.
\newblock Decoupled weight decay regularization.
\newblock In \emph{International Conference on Learning Representations}, 2019.
\newblock URL \url{https://openreview.net/forum?id=Bkg6RiCqY7}.

\bibitem[Luce(1959)]{Luce:1959choice}
R.~Duncan Luce.
\newblock \emph{Individual Choice Behavior: A Theoretical Analysis}.
\newblock John Wiley \& Sons, New York, 1959.

\bibitem[Plackett(1975)]{Plackett:1975permutations}
Robin~L. Plackett.
\newblock The analysis of permutations.
\newblock \emph{Applied Statistics}, 24\penalty0 (2):\penalty0 193--202, 1975.
\newblock \doi{10.2307/2346567}.

\end{thebibliography}
\endgroup

\end{document}